\documentclass{article}

\usepackage{PRIMEarxiv}

\usepackage[utf8]{inputenc} % allow utf-8 input
\usepackage[T1]{fontenc}    % use 8-bit T1 fonts
\usepackage{url}            % simple URL typesetting
\usepackage{booktabs}       % professional-quality tables
\usepackage{amsfonts}       % blackboard math symbols
\usepackage{nicefrac}       % compact symbols for 1/2, etc.
\usepackage{graphicx}
\usepackage{breakcites}		%Kai: Also Possible for breaking cites
\usepackage[square,sort,comma,numbers]{natbib}
\usepackage{float}
\usepackage{subfig}

\usepackage{microtype}      % microtypography
\usepackage{lipsum}
\usepackage{fancyhdr}       % header
\usepackage{graphicx}       % graphics
\graphicspath{{media/}}     % organize your images and other figures under media/ folder

\usepackage{latexsym}
\usepackage{times}
\usepackage{epsfig}
\usepackage{amsmath}
\usepackage{amssymb}
\usepackage{booktabs} % Kai: professional tables
\usepackage{multirow} % Kai: Used for UCI-Table
\usepackage{multicol} % Kai: Used for UCI-Table
\usepackage[pagebackref=true,breaklinks=true,colorlinks,bookmarks=false]{hyperref}

\title{Single-shot Bayesian approximation for neural networks
	\thanks{\textit{\underline{Citation}}: 
		\textbf{Authors. Title. Pages.... DOI:000000/11111.}} 
}

\DeclareMathOperator{\sm}{sm}
\DeclareMathOperator{\Cat}{Cat}
\renewcommand{\cite}[1]{{\citep {#1}}}

\title{Single-shot Bayesian approximation for neural networks%\thanks{Grants or other notes
}
\author{
	Kai Brach \\
	Elektrobit Automotive GmbH\\
	Germany\\
	\texttt{kai.brach@elektrobit.com}           % \\
	\And
	Beate Sick \\
	IDP, Zurich University of Applied Sciences \\ 
	EBPI, University of Zurich\\
	Switzerland\\
	\texttt{sick@zhaw.ch}           % \\
	\And
	Oliver D{\"u}rr \\
	IOS, Konstanz University of Applied Sciences\\
	Germany\\
	\texttt{oliver.duerr@htwg-konstanz.de} % \\
}

\begin{document}
	\maketitle
	
	\begin{abstract}
Deep neural networks (NNs) are known for their high-prediction performances. However, NNs are prone to yield unreliable predictions when encountering completely new situations without indicating their uncertainty. Bayesian variants of NNs (BNNs), such as Monte Carlo (MC) dropout BNNs, do provide uncertainty measures and simultaneously increase the prediction performance. The only disadvantage of BNNs is their higher computation time during test time because they rely on a sampling approach. Here we present a single-shot MC dropout approximation that preserves the advantages of BNNs while being as fast as NNs. Our approach is based on moment propagation (MP) and allows to analytically approximate the expected value and the variance of the MC dropout signal for commonly used layers in NNs, i.e. convolution, max pooling, dense, softmax, and dropout layers. The MP approach can convert an NN into a BNN without re-training given the NN has been trained with standard dropout. We evaluate our approach on different benchmark datasets and a simulated toy example in a classification and regression setting. We demonstrate that our single-shot MC dropout approximation resembles the point estimate and the uncertainty estimate of the predictive distribution that is achieved with an MC approach, while being fast enough for real-time deployments of BNNs. We show that using part of the saved time to combine our MP approach with deep ensemble techniques does further improve the uncertainty measures.
\end{abstract}
	
	
	%%%%%%%%% BODY TEXT
\section{Introduction}
\label{sec:intro}

Deep neural networks (NNs) have arisen as the dominant technique in many fields of computer vision in the last decade for the analysis of perceptual data.

NNs are also proposed for safety-critical applications such as autonomous driving \cite{Bojarski2016,  Grigorescu2019} or medical applications \cite{dolz2018}. However, classical NNs have deficits in capturing the model uncertainty \cite{Kendall2017,Gal2016}. But for safety-critical applications, it is mandatory to provide an uncertainty measure that can be used to identify unreliable predictions \cite{Michelmore2018,Feng2018,Harakeh2019,Miller2018,McAllister2017}. These can be situations that are completely different from all that occurred during training. In many application areas such as robotics or autonomous driving \cite{Sunderhauf2018, Bojarski2016}, where machines interact with humans, it is furthermore important to identify those situations in real time where a model prediction is unreliable and a human intervention is necessary.

Bayesian NNs (BNNs) \cite{MacKay1992} are an established method to compute an uncertainty measure for individual model predictions that take the model uncertainty into account, in addition to the uncertainty inherent in the data. However, state-of-the-art BNNs require sampling during deployment.
This leads to enlarged computation times compared to classical NNs limiting their usage for real-time applications. This work overcomes this drawback by providing a method that allows to approximate the expected value and variance of a BNN’s predictive distribution in a single run. It has therefore a comparable computation time as a classical NN. We focus here on a sampling free approximation for a special variant of BNNs which is known as MC dropout  \cite{Gal2016}. 

Ensembling-based models take an alternative approach to estimate uncertainties and have been successfully applied to NNs \cite{lak2017,pearce2020}. Recently it has been shown that ensembles of BNNs can further improve the quality of the uncertainty measure of both approaches  \cite{Wilson2020}. 
	\section{Related work}
\subsection{Quantization of uncertainty}
For the discussion of uncertainty, it is beneficial to treat deep learning (DL) models in a probabilistic framework where distributions are used to capture uncertainties. It's common to distinguish between aleatoric uncertainty capturing the data inherent variability and epistemic uncertainty capturing the uncertainty of the model parameter.

Aleatoric uncertainty can be modeled by controlling the parameters of a predictive distribution $p(y|x)$ of the outcome $y$ given $x$ via the output nodes of a NN, e.g. for a Gaussian predictive distribution $p(y|x)=N(y;\mu, \sigma)$ $\mu$ and $\sigma$ need to be controlled by an NN. For more complex predictive distributions for which the distribution family cannot be specified, a parametric transformation model can be used \cite{sick2020}. In computer vision, the input $x$ is often an image and outcome $y$ may be a scalar (regression), a categorical variable (classification), or might have as many categorical variables as there are pixels (semantic segmentation). These models are trained on the training data $D=(x_i,y_i)_{i=1}^N$ minimizing the negative log-likelihood 

\begin{equation}
  \mathrm{NLL} = -\frac{1}{N} \sum_{i=1}^N \log\left(p(y_i|x_i)\right),
  \label{eq:nll}
\end{equation}

To evaluate probabilistic models, the NLL on the test data is used. It can be shown that the NLL is a strictly proper scoring rule \cite{Gneiting2007}, meaning it gets only minimal if, and only if, the predictive distribution corresponds to the data generation distribution. The model with the smallest test NLL yields the most appropriate predictive distribution. 

For regression type problems a common choice for the family of the predictive distribution is the normal distribution $N(\mu(x), \sigma(x)^2)$, where the variance $\sigma(x)^2$ quantifies the uncertainty at a particular value of $x$. Several choices exist to model the variance $\sigma^2(x)$. In \cite{Gal2016} the variance is treated as the reciprocal of a hyperparameter $\tau$, $\sigma^2=\tau^{-1}$, and the value $\tau$ is chosen by cross-validation. Alternatively, the NN can be used to learn the variance from the data, if necessary in dependence on the input $x$ \cite{Kendall2017}. Note that a non-probabilistic regression model that is trained with the mean-square-error (MSE) loss corresponds to a model with a Gaussian predictive distribution and a constant variance.    

For classification, a common choice for the predictive distribution is a categorical distribution $p(y|x) = \Cat(y;{\mathbb\pi}(x))$ where the number of classes defines the length of the parameter vector ${\mathbb\pi}(x)$ with ${\pi}_i(x)$ being the predicted probability of class $i$. Again, the parameters of the predictive distribution are controlled by an NN that computes ${\mathbf\pi}(x)$ by turning the NN output logits of the last layer into probabilities using the softmax function. 

Another, less common approach, is to use a Dirichlet distribution instead of the categorical \cite{Gast2018}. Alternatively, an NN can be used to control the parameter of a transformation model to estimate the predictive distributions of an disordered or ordered categorical outcome \cite{kook2020}. 

From the predictive distribution an uncertainty measure can be constructed. For regression type problems the spread of the distribution, quantified by the variance $\sigma^2(x)$, can be used. For classification tasks it is less obvious which measure captures the spread of the probabilities across the classes best. A common measure is the entropy: 
\begin{equation}
   H(x) = - \sum_i^K \pi_i(x) \cdot \ln(\pi_i(x)),
   \label{eq:entropy}
\end{equation}
The entropy takes its minimal value zero, if the whole probability mass of one is assigned to only a single class indicating that there is no uncertainty about the outcome. Another common choice for the uncertainty of a prediction is $1-\max(\pi_i)$ \cite{Gal2017}. %Again, if the whole probability mass of one is assigned to a single class, no uncertainty is left to the prediction. 

So far, the spread of the predictive distribution reflects the fact that the input $x$ does not completely define the value of the outcome $y$. This corresponds to the uncertainty inherent in the data (aleatoric uncertainty). Note that the aleatoric uncertainty cannot be reduced when adding more training data. However, it is obvious that a lack of training data hinders a precise estimation of the model parameters, leading to a high uncertainty. This kind of model uncertainty is called epistemic uncertainty. This gets especially important when leaving the range of the training data in case of regression, or when presenting an instance of a novel class in case of classification. 

Recent work shows that ensembles of NNs can be used to capture epistemic uncertainty. In this approach an ensemble of NNs with identical architecture are trained on the same data but with different random initialization of the weights \cite{lak2017, Wilson2020}. During test time the differently initialized NNs of the ensemble yield for the same input different outputs, from which the moments of the predictive distribution can be estimated.

A well-established approach to model the epistemic uncertainty of a model is Bayesian modeling. In BNNs the fixed weights are replaced by distributions $p(w|D)$. The more training data are available, the narrower these posteriors get indicating a low model uncertainty. The uncertainty of the weights translates in the uncertainty of the predictive distribution. The predictive distribution in a BNN is given by marginalizing over different weight configurations via:  
\begin{equation}
    p(y|x) = \int p(y|x,w) p(w|D) \; dw
\label{eq:bay_cpd}
\end{equation}

An analytical solution for (\ref{eq:bay_cpd}) can not be determined for a BNN with millions of weights. For small networks it is possible to do a Markov Chain Monte Carlo (MCMC) simulation \cite{gustafsson2020}. In the MCMC $T$ samples of $w_t \sim p(w|D)$ from the posterior are drawn. Using these samples, the integration in (\ref{eq:bay_cpd}) can be approximated by summation:

\begin{equation}
    p(y|x) \approx \frac{1}{T} \sum_{t=1}^T p(y|x,w_t)
\label{eq:bayes_sample}
\end{equation}

For common-sized networks MCMC is way too slow and approximations of the posterior need to be used. One way to approximate the posterior is the variational inference approach, where a simple parametric distribution $q_\phi(w)$, called variational distribution, is used to approximate the true posterior. The variational parameter $\phi$ of the variational distribution $q_\phi(w)$ is tuned to minimize the Kullback–Leibler divergence between the true posterior and  $q_\phi(w)$. A common approach is the mean-field variational approximation, in which the weight distributions are modeled independently, often using simple independent Gaussians for $q_\phi(w)$ \cite{Kingma2014,Blundell2015}. While complex multivariate distributions are possible in principle \cite{louizos2017}, recent work \cite{farquhar2020} suggests that for deep neural networks the mean-field assumption gives a sufficient approximation to the predictive distribution in equation \ref{eq:bay_cpd}. 
The mean-field variational approximation can be further approximated, which is commonly done by MC dropout. 

\subsection{MC dropout}
In case of MC dropout BNNs, each weight distribution is a Bernoulli distribution: the weight takes with the dropout probability $p^*$ the value zero and with probability $1-p^*$ the value $w$. All weights starting from the same neuron are set to zero simultaneously. The dropout probability $p^*$ is usually treated as a fixed hyperparameter and the weight-value $w$ is the variational parameter that is tuned during training. 
In contrast to standard dropout \cite{Srivastava2014}, the weights in MC dropout are not frozen and rescaled after training, but the dropout procedure is also done during test time. It can be shown that MC dropout can be interpreted as a variational inference approach \cite{Gal2016}. 
MC dropout BNNs were successfully used in many applications, have proven to yield improved prediction performance, and allow to define uncertainty measures to identify individual unreliable predictions \cite{Ryu2019,Duerr2018,Kwon2020, herzog2020}. To employ a trained BNN in practice one performs $T$ runs of predictions $p(y|x,w_t)$ with different weight configurations $w_t \sim q_\phi$ during test time. As in the MCMC case, we can make use of equation \ref{eq:bayes_sample} to calculate  $p(y|x)$. In this way, the outcome distribution incorporates the epistemic and aleatoric uncertainty. 

The only drawback of an MC dropout BNN compared to its classical NN variant is the increased computing time due to the sampling procedure. The sampling procedure leads to a computing time that is prohibitive for many real-time applications like such as autonomous driving. To overcome this limitation, a single shot method, which does not require sampling, is desirable.

\subsection{Moment propagation}
\label{subsec:related_work_ep}

In this work, we are aiming to approximate MC dropout BNN without sacrificing their benefits over NN, i.e. improved prediction performance and quantification of the epistemic uncertainty without the need to do time-costly sampling. To do so we do not treat the activation of the neural network as a distribution. This is similar to natural parameter networks \cite{Wang2016}, which are based on exponential distributions and so requires that the network components approximately keep the distributions in the exponential family. This disallows the use of important but highly non-linear network components such as softmax needed for classification or dropout needed for quantification of uncertainty. 

In contrast, our method mainly relies on statistical moment propagation (MP). More specifically, we propagate the expectation $E$ and the variance $V$ of the signal distribution, through the different layers of an NN. This approach is known in statistics as the delta method \cite{Dorfman1938}. In MP, we approximate the layer-wise transformations of the variance and the expected value of our signal. A similar approach has also been used for NNs before \cite{Frey1998, adachi2019, Jin2015, Gast2018, Loquercio2020, Brach2020}. 

However, quantifying the epistemic uncertainty \cite{Gast2018} and \cite{Loquercio2020} did not include the MC dropout activity in their framework, but did time-consuming MC dropout runs.  Due to our best knowledge, so far only \cite{postels2019} and \cite{Brach2020} have modeled the MC dropout within the MP framework thus allowing to evaluate the epistemic uncertainty in a single-shot approximation without the need to do several MC dropout runs. While \cite{postels2019} included the MC dropout in a sampling-free manner to estimate the ”injected” noise, they did not consider mean value but just propagated the variance. However, treating the mean value as part of the MP increases the predictive performance. Further, we here treat a more complete set of activations in contrast to our preliminary previous work \cite{Brach2020} for regression-like problems, by including convolutions, max pooling, and softmax which are important for classification problems.
	\section{Methods}
\label{sec:methods}
\begin{figure*}[hbt!]
    \centering
    \includegraphics[width=1.\textwidth]{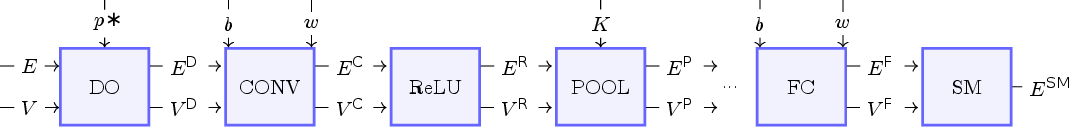}
    \caption{Overview of the proposed method. The expectation E and variance V flow through different layers of the network in a single forward pass. Shown is an example configuration in which dropout (DO) is followed by a convolution (CONV), an ReLU activation and a max pooling layer (POOL). The network output is specified by a softmax activation (SM) which follows a dense layer (FC) with linear activation. More complex networks can be built by different arrangements of the individual blocks. }
    \label{fig:overview_cnn}
\end{figure*}

The goal of our MP method is to approximate the expected value E and the variance V of the predicted output that is obtained by the above described MC dropout method. When propagating the signal of an observation through an MC dropout network, we get each layer with $p$ nodes, an activation signal with an expected value $E$ (of dimension $p$) and a variance given by a variance-covariance matrix $V$ (of dimension $p \times p$). In the dropout layer, uncertainty is introduced yielding to a non-zero variance in intermediate layers and the output, despite the fact that the input signal variance is assumed to be zero. We neglect the effect of correlations between different activations, which are small anyway in deeper layers due to the decorrelation effect of the dropout. Hence, we only consider diagonal terms in the correlation matrix. In the following, we describe the essential layer types in fully connected and convolutional networks and how the expected value E and its variance V is propagated. As layer types we consider dropout, convolution, dense, max pooling, softmax, and ReLU activation layers. Figure \ref{fig:overview_cnn} provides an overview of the layer-wise abstraction.

\subsection{Dropout layer}
We start our discussion with the effect of MC dropout. Let $E_i$ be the expectation at the ith node of the input layer and $V_i$ the variance at the ith node. In a dropout layer the random value of a node $i$ is multiplied independently with a Bernoulli variable $Y \sim \tt{Bern(p^*)}$ that is either zero or one. The expectation $E_i^D$ of the ith node after dropout is then given by:

\begin{equation}
E^D_i = E_i(1-p^*)
\label{eq:e_do}    
\end{equation}

For computing the variance $V_i^D$ of the ith node after dropout, we use the fact that the variance $V(X \cdot Y)$ of the product of two independent random variables $X$ and $Y$, is given by \cite{Goodman1960}:

\begin{equation}
    V(X \cdot Y) = V(X) V(Y) + V(X) E^2(Y) + E^2(X) V(Y)
\end{equation}

With $V(Y)=p^*(1-p^*)$, we get:
\begin{equation}
V_i^D=V_i \cdot p^*(1-p^*)+V_i (1-p^*)^2+E^2_i \cdot p^*(1-p^*)
\label{eq:v_do}    
\end{equation}

Dropout is the only layer in our approach where uncertainty is created, even if the input has $V_i = 0$  and the output of the dropout layer has for $p^* \ne 0$ a variance $V_i^D > 0$.

\subsection{Dense and convolutional layer}
For the dense layer with $p$ input and $q$ output nodes, we compute the value of the ith output node  as $\sum_j^p w_{ji} x_j + b_i$, where $x_j, j=1 \dots p$ are the values of the input nodes. Using the linearity of the expectation, we get the expectation $E^F_i$ of the ith output node from the expectations, $E^F_j, j=1 \dots p$, of the input nodes:

\begin{equation}
E^F_i =\sum_{j=1}^p w_{ji} E_j + b_i
\label{eq:e_fc}    
\end{equation}

To calculate the change of the variance, we use the fact that the variance under a linear transformation behaves like $V( w_{ji} \cdot x_j + b)= w_{ji}^2 V(x_j)$. Further, we assume independence of the j different summands, yielding:  

\begin{equation}
V_i^F = \sum_{j=1}^p w^2_{ji} V_j
\label{eq:v_fc}    
\end{equation}

For convolutional layers, which are also just another form of affine linear layers, the results above stay true. The sum in equation (\ref{eq:e_fc}) is now over all neighboring pixels $j$ of $i$. For the expectation $E^C$, the usual convolutional formula stays valid and is now taken with respect to the expectation of the activations $E(x_j)$. In the technical implementation, the highly optimized convolution of the underlying deep learning framework can be used, one just has to provide the expectations E instead of the activations. For the variance $V_i^C$ the weights of the kernel have to be squared and a version without bias has to be used.

\subsection{ReLU activation layer}
To calculate the expectation $E^{R}_i$  and variance $V_i^R$ of the ith node after an ReLU, as a function of the $E_i$ and $V_i$ of the layer before the ReLU, we need to make a distributional assumption.  We assume that the input is standard Gaussian distributed, with $\phi(x)=N(x;0,1)$ the PDF, and $\Phi(x)$ the corresponding CDF, we get the following output  for the expectation and variance. Derivation is given by \cite{Frey1998}:

\begin{equation}
E^R_i = E_i \cdot \Phi\left(\frac{E_i}{\sqrt{V_i}}\right) + \sqrt{V_i} \cdot \phi\left(\frac{E_i}{\sqrt{V_i}}\right) 
\label{eq:e_relu}    
\end{equation}

\begin{equation}
% V^R_i = (\mu^2 + \sigma^2) \Phi\left(\frac{\mu}{\sigma}\right) + \mu \sigma \cdot \phi\left(\frac{\mu}{\sigma}\right) - (E^R_i)^2 
V^R_i = (E_i^2 + V_i) \Phi\left(\frac{E_i}{\sqrt{V_i}}\right) + E_i \sqrt{V_i} \cdot \phi\left(\frac{E_i}{\sqrt{V_i}}\right) - {E^R_i}^2
\label{eq:v_relu}    
\end{equation}

\subsection{Max pooling layer}
The most commonly used pooling operation in convolutional neural networks (CNN) is max pooling. It has found to be superior for capturing invariances in image-like data, compared to other sub-sampling operations \cite{Suarez2018,Scherer2010}. Max pooling takes the maximum values out of each pooling-region, specified by a $K=(N \times N)$ kernel matrix and creates a new matrix containing the maximum values.

 For MP, we again have to make distributional assumptions. Specifically, we assume that the values within each pooling-region follow independent Gaussians with $ x_k \sim N(E_k, V_k)$, where $k=1,\dots, K$.

Extracting the maximum values of the $K$-independent Gaussians cannot be applied directly using the conventional max pooling approach. To our best knowledge, there exists no closed-form expression to calculate the maximum of more than two Gaussian variables directly. We therefore decompose the problem into repeated calculations of two Gaussians. The exact moments  $E$ and $V$ of $X = \max(X_1,X_2)$ for two random variables $X_i \sim N(E_i,V_i)$, $i \in {1,2}$ are given by  \cite{Nadarajah2008} as 

\begin{equation}
\begin{split}
E^P(X) = E_1 \Phi\left(\frac{E_1 - E_2}{\theta}\right) + E_2\Phi\left(\frac{E_2 - E_1}{\theta}\right) \\
+ \theta \phi \left(\frac{E_2 - E_1}{\theta}\right)
\end{split}
\label{eq:e_maxpool}    
\end{equation}

\begin{equation}
\begin{split}
V^P(X) = \left( V_1 + E^2_1\right)\Phi\left(\frac{E_1 - E_2}{\theta}\right) \\
+\left( V_2 + E^2_2\right)\Phi\left(\frac{E_2 - E_1}{\theta}\right) \\
+\left( E_1 + E_2\right)\theta \phi \left(\frac{E_1 - E_2}{\theta}\right)
\end{split}
\label{eq:v_maxpool}        
\end{equation}

where $\theta = \sqrt{V_1+V_2}$ for uncorrelated random variables.

To calculate the maximum of an arbitrary number of $K$ random variables $X=\max(X_1,X_2,X_3,\dots,X_k)$ for $k=1 \dots K$, we apply the equations (\ref{eq:e_maxpool}) and (\ref{eq:v_maxpool}) $K-1$ times. 
For example a $K=(2 \times 2$) kernel matrix lead to a vector $X = (X_1,\dots,X_4)$ of four independent Gaussians with $X_k \sim N(E_k,V_k)$ where $k=1,\dots, 4$. To find the maximum of $X$, we need to calculate the maximum of two independent random variables recursively.  

In \cite{Jin2015} the random variables $X_k$ have been sorted using a heuristic. However, in preliminary studies we did not find a consistent beneficial effect of the ordering and did refrain from reordering.

\subsection{Softmax layer}
For classification, the output of the NN are the probabilities $\pi_i$, for the respective classes $k=1,\ldots K$. Usually the probabilities are calculated via a softmax ($\sm$) layer, from the output $z_i$ (logits) of the final layer of a network as

\begin{equation}
    \pi_i = \sm(z_i) = \frac{\exp(z_i)}{\sum_{k=1}^K \exp(z_k)}
\end{equation}

Again, we assume that the input $z_i$ follows independent Gaussians $z_i \sim N(E_i, V_i)$. An approximation, for the expectation of the softmax, is given in \cite{Daunizeau2017} with $\sigma(z)$ being the sigmoid function: 

\begin{equation}
    E(\sm(z_i)) \approx \frac{1}{2-K+\sum_{k' \ne k} \frac{1}{E(\sigma(z_i - z_k'))}}
\label{eq:sm1}
\end{equation}

Since $z_i$ and $z_k'$ are independent Gaussian $z_k - z_k' \sim N(E(z_k)-E(z_k'), V(Z_k)+V(z_k'))$. For the calculation of the equation (\ref{eq:sm1}), we thus need the expectation of a Gaussian "piped through" a sigmoid function. While \cite{Daunizeau2017} also gives an approximation for this, we found in Monte Carlo studies that the approximation given in equation (8.68) of Murphy's book \cite{Murphy2012} yields better results. In this approximation the sigmoid function is approximated by the CDF $\Phi$ of the standard Gaussian, as: 

\begin{equation}
    \sigma(x) \approx \int_\infty^x N(\frac{\pi}{8} \cdot x';0,1) dx' = \Phi(\frac{\pi}{8} x)
\end{equation}

This allows to write the expectation $E[\sigma(x)]$ of a $X \sim N(E_i, V_i)$ (see \cite{Murphy2012}) as

\begin{equation}
    E_{x \sim N(E_i,V_i)}[\sigma(x)] = \Phi\left( \frac{E_i}{\sqrt{V_i + 8 / \pi}}\right)    
\end{equation}

TensorFlow \footnote {\url{https://www.tensorflow.org/}} implementations of all described layers are available on GitHub \footnote {\url{https://github.com/kaibrach/Moment-Propagation}}. Note that in order to use the MP approach to transfer a DNN into a BDNN no re-training has to be done, provided that the original network has been trained with standard dropout.
	\section{Results}
\label{sec:results}

\subsection{Toy dataset}
\begin{figure}[h]
    \centering
    \includegraphics[width=0.5\textwidth]
    {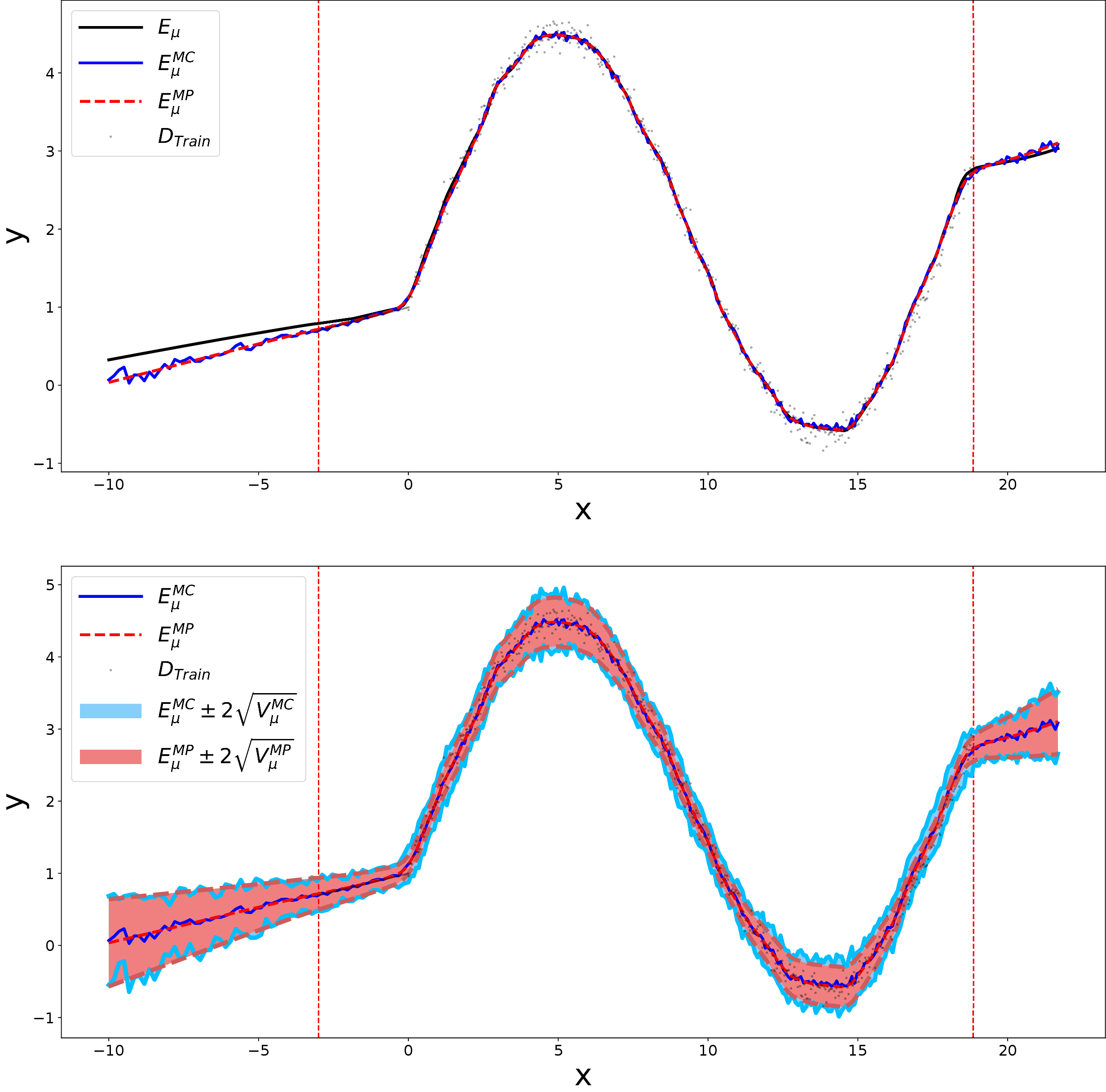}
    \caption{Comparison of the MP and MC dropout results. 
    The NNs were fitted on training data in the range of -3 to 19 (vertical lines). In the upper panel the estimated expectations of the MC, the MP, and additionally the NN methods are compared. In the lower panel the predicted spread of $\mu(t)$ for the MC and MP method is shown.}
    \label{fig:toy}
\end{figure}

We first apply our approach on a one dimensional regression toy dataset, with only one input feature and additional Gaussian noise of $\mu_{noise}=0$ and $\sigma_{noise}=0.1$ on training data. We use a fully connected NN with three layers each with 256 nodes, ReLU activations, and dropout after the dense layers. We have a single node in the output layer which is interpreted as the expected value $\mu$ of the conditional outcome distribution $p(y|x)$. We train the network for 2000 epochs using the MSE loss and apply dropout with $p^*=0.3$. From the MC dropout BNN, we get at each x-position $T=30$ MC samples $\mu_t(x)$ from which we can estimate the expectation $E_\mu$ by the mean and $V_\mu$ by the variance of $\mu_t(x)$. We use our MP approach on the same data to approximate the expected value $E_\mu$ and the variance $V_\mu$ of $\mu$ at each $x$ position. In addition we use an NN in which dropout has only been used during training yielding a deterministic output $\mu(x)$. All three approaches yield  within the range of the training data nearly identical results for the estimated $\mu(x)$ (see upper panel in figure \ref{fig:toy}). We attribute this to the fact that we have plenty of training data and so the epistemic uncertainty is neglectable. In the lower panel of figure \ref{fig:toy} a comparison of the uncertainty of $\mu(x)$ is shown by displaying an interval given by the expected value of $\mu(x)$ plus/minus two times the standard deviation of $\mu(x)$. Here the width of the resulting intervals of a BNN via the MP approach and the MC dropout are comparable (the NN does not yield an estimation for the epistemic uncertainty). This demonstrates that our MP approach is equally useful as the MC approach to provide epistemic uncertainty estimations.  

% UCI Dataset Evaluation
\begin{table*}[t]
	\caption{Comparison of the average prediction performance in test RMSE (root-mean-square error), test NLL (negative log-likelihood) and test RT (runtime) including $\pm$ standard error on UCI regression benchmark datasets between MC and MP. \textit{N} and \textit{Q} correspond to the dataset size and the input dimension. For all test measures, smaller means better.}
	\label{tbl:uci_benchmark}
	\vskip 0.15in
	\begin{center}
		\begin{small}
			\begin{sc}
				\resizebox{\linewidth}{!}{\begin{tabular}{lllcccccc}
						\toprule
						\multirow{2}{*}{Dataset} &
						\multirow{2}{*}{\textit{N}} &
						\multirow{2}{*}{\textit{Q}} &
						\multicolumn{2}{c}{Test RMSE} &
						\multicolumn{2}{c}{Test NLL} &
						\multicolumn{2}{c}{Test RT [s]} \\
						& & & {MC} & {MP} & {MC} & {MP} & {MC} & {MP} \\
						\midrule
	                    Boston		&506	&13 &3.14 $\pm{0.20}$	&3.10 $\pm{0.20}$	&2.57 $\pm{0.07}$	&2.56 $\pm{0.08}$	&2.51 $\pm{0.03}$	&0.04 $\pm{0.00}$\\
						Concrete	&1,030	&8 	&5.46 $\pm{0.12}$	&5.40 $\pm{0.12}$	&3.12 $\pm{0.02}$	&3.13 $\pm{0.03}$	&3.37 $\pm{0.04}$	&0.04 $\pm{0.00}$\\	
						Energy		&768	&8 	&1.65 $\pm{0.05}$	&1.61 $\pm{0.05}$	&1.95 $\pm{0.04}$	&2.01 $\pm{0.04}$	&2.84 $\pm{0.03}$	&0.04 $\pm{0.00}$\\	
						Kin8nm		&8,192	&8	&0.08 $\pm{0.00}$	&0.08 $\pm{0.00}$	&-1.10 $\pm{0.01}$	&-1.11 $\pm{0.01}$	&7.37 $\pm{0.06}$	&0.04 $\pm{0.00}$\\	
						Naval		&11,934	&16 &0.00 $\pm{0.00}$	&0.00 $\pm{0.00}$	&-4.36 $\pm{0.01}$	&-3.60 $\pm{0.01}$	&9.69 $\pm{0.11}$	&0.04 $\pm{0.00}$\\
						Power		&9,568	&4 	&4.05 $\pm{0.04}$	&4.04 $\pm{0.04}$	&2.82 $\pm{0.01}$	&2.84 $\pm{0.01}$	&6.85 $\pm{0.07}$	&0.04 $\pm{0.00}$\\		
						Protein		&45,730	&9  &4.42 $\pm{0.03}$	&4.41 $\pm{0.02}$	&2.90 $\pm{0.00}$	&2.91 $\pm{0.00}$	&31.38 $\pm{0.09}$	&0.05 $\pm{0.00}$\\
						Wine		&1,599	&11 &0.63 $\pm{0.01}$	&0.63 $\pm{0.01}$	&0.95 $\pm{0.01}$	&0.95 $\pm{0.01}$	&4.78 $\pm{0.01}$	&0.04 $\pm{0.00}$\\		
						Yacht		&308	&6 	&2.93 $\pm{0.22}$	&2.91 $\pm{0.26}$	&2.35 $\pm{0.07}$	&2.11 $\pm{0.07}$	&2.01 $\pm{0.01}$	&0.04 $\pm{0.00}$\\
						\bottomrule
				\end{tabular}}
			\end{sc}
		\end{small}
	\end{center}
	\vskip -0.1in
\end{table*}

\subsection{UCI-datasets}
To benchmark our method in regression tasks, we redo the analysis of \cite{Gal2016} for the UCI regression benchmark dataset, using the same NN model structure as provided in the experiment. The NN model is a fully connected neural network including one hidden layer with ReLU activation in which the predictive distribution $p(y|x)$ is estimated via:

\begin{equation}
p(y|x) = \frac{1}{T} \sum_t N(y ; \mu_t(x), \tau^{-1})
\label{eq:gal_cpd}    
\end{equation}

Again $\mu_t(x)$ is the single output of the BNN for the tth out of $T=10\;000$ MC runs. To derive a predictive distribution, Gal assumes in each run a Gaussian distribution, centered at $\mu$, and a precision $\tau$, corresponding to the reciprocal of the variance. The parameter $\mu$ is received from the NN and $\tau$ is treated as a hyperparameter. For the MP model, the MC sampling in equation (\ref{eq:gal_cpd}) is replaced by integration: 

\begin{eqnarray}
p(y|x) &=& \int N(y ; \mu', \tau^{-1}) N(\mu'; E\textsuperscript{MP}, V\textsuperscript{MP}) \; d\mu'\nonumber\\ 
&=& N(y;E\textsuperscript{MP}, V\textsuperscript{MP} + \tau^{-1})
\label{eq:mp_cpd}    
\end{eqnarray}

For comparison of our MP method against MC dropout we follow the same protocol as \cite{Gal2016}\footnote {\url{https://github.com/yaringal/DropoutUncertaintyExps}}. Accordingly, we train the network for 10$\times$ the epochs provided in the individual dataset configuration. As described in \cite{Gal2016}, an excessive  grid search over the dropout rate $p^*=0.005,0.01,0.05,0.1$, and different values of the precision $\tau$ is done. The hyperparameters minimizing the validation NLL are chosen and applied on the test set. 

We report in table \ref{tbl:uci_benchmark} the test performance (RMSE and NLL) achieved via MC BNN using the optimal hyperparameters  for the different UCI datasets. We also report the test RMSE and the NLL achieved with our MP method. Allover, the MC and MP approaches  produces very similar results (the NAVAL dataset is a special case, because the outcome depends in a deterministic manner on the input, and also the epistemic uncertainty is close to zero). But we want to stress that our MP approach is much faster, as shown in the last column in the table \ref{tbl:uci_benchmark}, because it has only to perform one forward pass instead of $T=10\;000$ forward passes.

\subsection{Classification}

We also want to benchmark our MP method on a classification task. For this we use a conventional CNN as illustrated in figure \ref{fig:overview_cnn} on CIFAR-10 data\footnote {\url{https://www.cs.toronto.edu/~kriz/cifar.html}}. We set up a CNN with three convolutional layers with filter sizes of 16, 32, and 64, and 3x3 kernel.
ReLU activations followed by a max pooling operation are applied after each convolution. Dropout with $p^*=0.3$ is applied to each max-pooling layer as well as to the two additional dense layers with 128 neurons each. In the last layer of the CNN a softmax function is used to transform the logits $z_i$ into a vector of probabilities  $(\pi_1, \dots, \pi_K)$ for the categorical predictive distribution with $K$ classes. 

To investigate if the MP approach approximates the MC dropout uncertainty we performed an out-of-distribution (OOD) experiment. This experiment is set up by training the CNN only on five out of ten classes of the CIFAR-10 dataset, which we call in-distribution (IND) classes (airplane, cat, deer, dog, ship); the remaining classes are OOD classes (automobile, bird, frog, horse, truck). This allows us to evaluate the quality of the uncertainty measures for aleatoric (for the IND setting) and combined aleatoric and epistemic (for the OOD setting) uncertainty. We used a train/validation split of $80/20$ and as loss function the NLL of the categorical distribution $p(y|x) = \Cat({\pi})$. 

We apply an automatic reduction of the learning rate after five epochs by a factor of 15\% if the monitored validation NLL does not improve. Further, we use early stopping mechanism if no decrease in the validation NLL for a consecutive number of ten epochs is observed. We use the same trained network during test time in three settings, as a non-Bayesian NN with fixed weights (NN), an MC dropout version with $T=50$ samples (MC), and our approximation (MP). 

\begin{figure}[ht]
    \centering
    \includegraphics[width=0.48\textwidth]
    {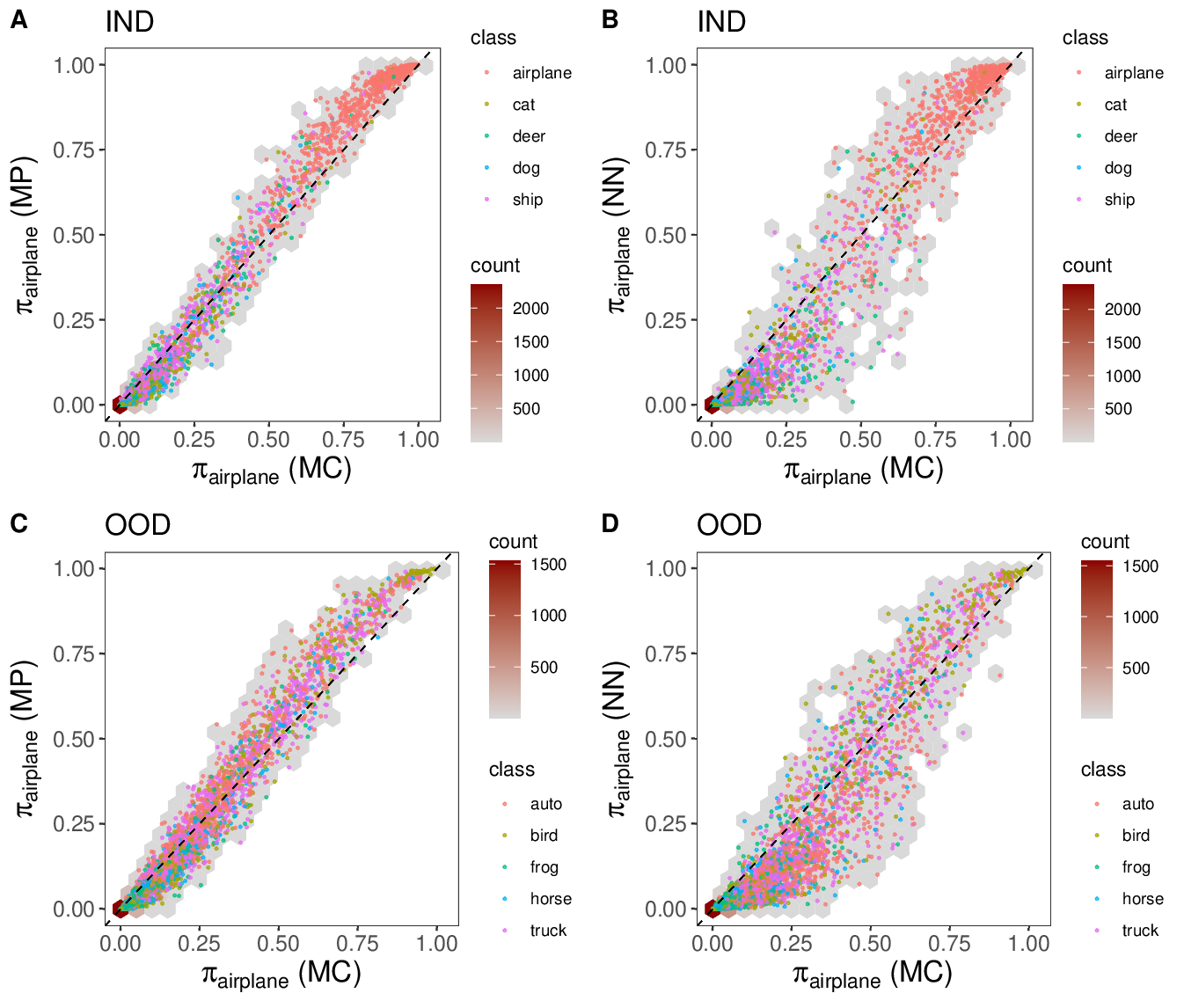}
    \caption{Density plots for comparison of the marginal probability values that were predicted for the IND class airplane in the OOD experiment, where the true class is indicated by the color code. The upper row (A, B) shows the results for 5000 IND images. The lower row (C, D) shows results for 5000 OOD images. Shown is the the MC dropout approach against our approximation (MP) (panels A and C) and against a NN (panels B and D).}
    \label{fig:p}
\end{figure}

As a first test of our MP method, we compare how well our approach approximates the MC method with respect to the predictive distribution $p(y|x)=\Cat(\pi(x))$. A direct comparison of the output of the network, the five estimated parameters $\pi_i$, requires a multivariate treatment. Since the $\pi_i$s are probabilities the restriction $\sum_i \pi_i = 1$ applies. A sound discussion would need to take this into account and compare the probabilities in the framework of compositional data which is beyond the scope of this study. To still provide an impression of the similarity of the probabilities received with the three methods (NN, MC, MP), we compare the marginal probabilities of the first class (the IND class airplane) for all predictions in the test set (see figure \ref{fig:p}). 

In the upper panel of figure \ref{fig:p} we look only at samples from the test set that correspond to a class that is in the training set, and we can see from the color code that for airplane samples all models predict usually high airplane probabilities whereas for non-airplane samples all models predict usually rather low airplane probabilities. For the 5000 OOD images  (lower panel in figure \ref{fig:p}), by definition, none of the possible classes are correct, leading to a larger spread of the probability values, but still the wrongly assigned airplane probabilities are quite reproducible across all three models.
Comparing the left versus the right panel of figure \ref{fig:p} shows that the results from the MP and MC models are much closer to each other than the results from the NN and MC model, indicating again that our MP approach is a good approximation for the MC model.

\begin{figure}[ht]
    \centering
    \includegraphics[width=0.5\textwidth]
    {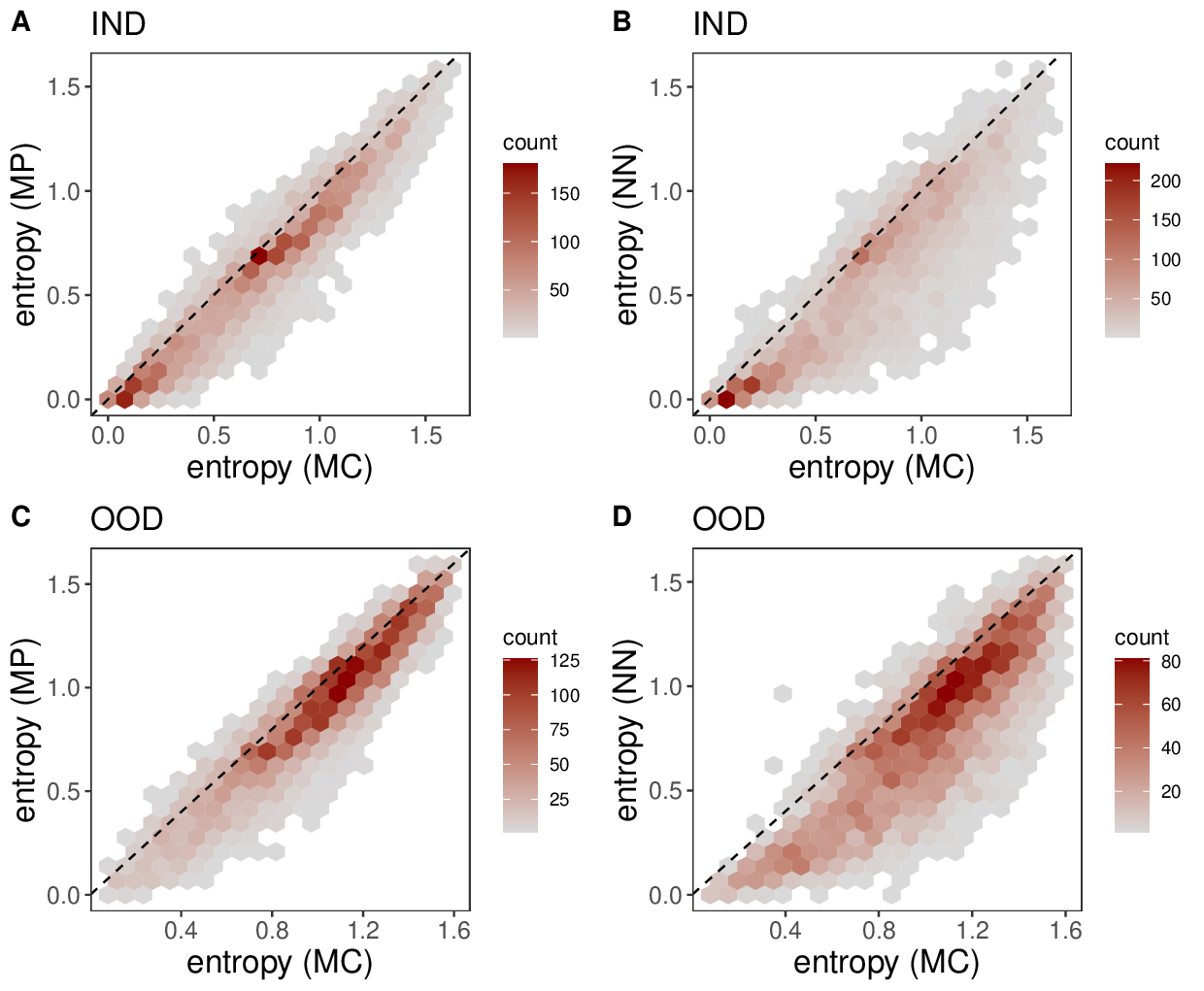}
    \caption{Density plots for comparison of entropy values of the predicted categorical CPDs in the OOD experiment. On the one hand the MC model is compared to the MP model (panels A and C) and on the other hand to the NN model (panels B and D). In the upper panel the comparison is done on 5000 IND images, and in the lower panel the comparison is done on 5000 OOD images .  
    }
    \label{fig:entropy}
\end{figure}

For a global comparison of the predictive distribution taking all classes into account, we use the entropy from equation (\ref{eq:entropy}) quantifying the uncertainty of the predictive distribution in one number.
For both test samples (IND and OOD), the entropy values from the MP method nicely approximate the entropy values from the MC dropout method (see left panel of figure \ref{fig:entropy}). Most data lies on a diagonal, and the Pearson correlation coefficient between MP and MC is 0.981 with a 95\% CI [0.980, 0.982] for the IND images, and 0.970  [0.968 0.971] for the out-train-classes images. Figure \ref{fig:entropy} shows also that the  non-Bayesian network (NN) has a significant weaker correlation with the MC approach, the correlation here is  0.931  [0.927 0.934] for the in training examples and  0.893  [0.887 0.899] for the novel examples.       

Figure \ref{fig:entropy} also shows that the predictive distribution of IND classes typically has a low entropy (indicating a low uncertainty of the predictions), and OOD classes often have larger entropy values (indicating a high uncertainty).

Altogether, our MP model yields a good approximation for the MC dropout model. MC dropout models are usually used for two reasons, first to quantify the uncertainty and secondly to enhance the predictive performance. We now check if our MP method approximates the MC dropout results with respect to both aspects. 

To access the predictive performance, we use all ten classes of the CIFAR-10 dataset for training and testing. The achieved predictive performance in terms of accuracy is 0.7168 with a 95\% Wilson CI [0.7079, 0.7255] in case of NNs, 0.7467 [0.7381 0.7551] in case of MC, and 0.7459 [0.7372 0.7543] in case of MP. In terms of the NLL, where a smaller value indicates a better performance, we see a similar picture and achieve $0.82 \pm 0.02$ (NN), $0.77 \pm 0.02$ (MC), and $0.75 \pm 0.02$ (MP)\footnote{The 95\% CI have been calculated using $\pm 1.96$ times the standard error}. For both measures the MP and MC significantly outperform the simple neural network, while no significant performance difference has been observed between the MC method and the fast MP approximation.  

To study if the quality of the uncertainty measures resulting from MC and MP are equally reliable, we want to check if the uncertainty measures from both methods perform equally good in identifying novel classes. For this purpose we switch back to the OOD experiment in which five classes were left out for training. We use the entropy as uncertainty-score to decide if an example is from a novel class \cite{Gal2017, Duerr2018}. 

\begin{figure}[ht]%
    \centering
        \begin{minipage}[c]{0.95\textwidth}
        \centering
            \includegraphics[width=0.5\textwidth]{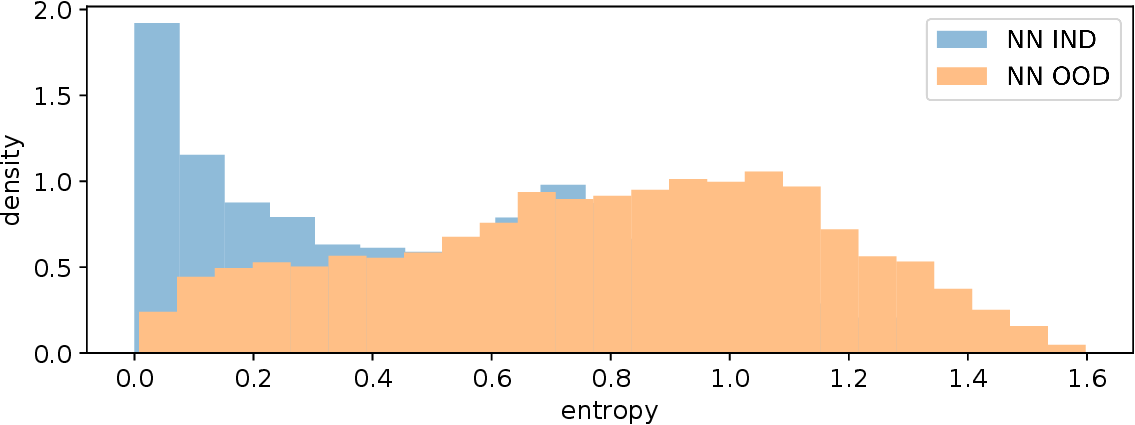}
        \end{minipage}
        
        \begin{minipage}[c]{0.95\textwidth}
        \centering
            \includegraphics[width=0.5\textwidth]{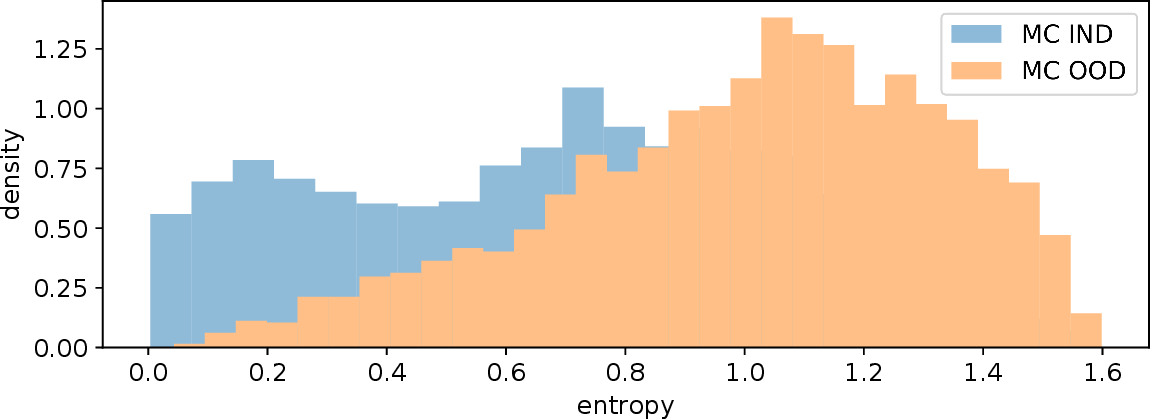}
        \end{minipage}
        
        \begin{minipage}[c]{0.95\textwidth}
        \centering
            \includegraphics[width=0.5\textwidth]{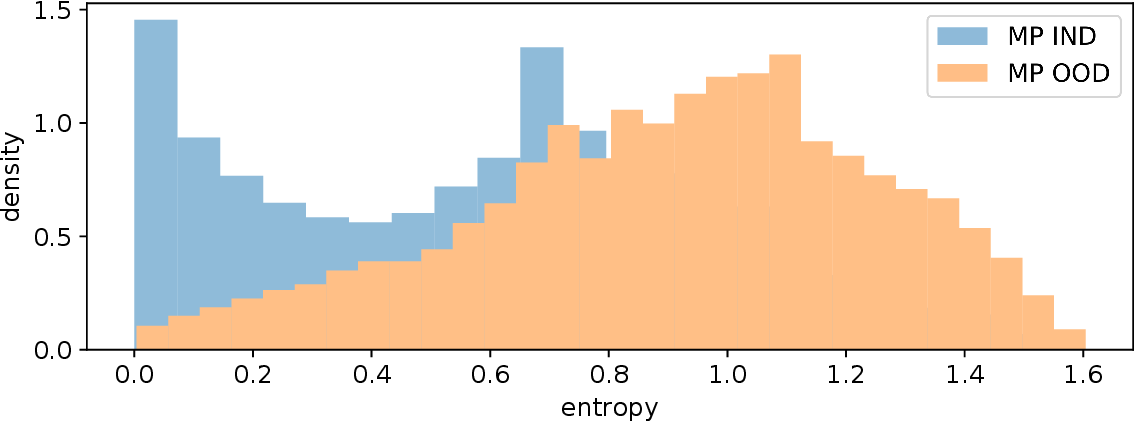}
        \end{minipage}
        
        \caption{\label{fig:ent_hist} Entropy as OOD score. Shown are the histograms of the entropy separated between IND and OOD for NN, MC, MP from top to bottom.}
\end{figure}

In the following, we present a ROC analysis for the ability to detect the OOD classes examples. We use the entropy to distinguish between in-distribution (IND) from OOD examples.  In figure \ref{fig:ent_hist} the histograms for (NN) upper panel, (MC) middle panel, and (MP) lower panel are shown. As expected the IND examples generally have a lower entropy. This effect is especially pronounced for the MP and MC case.  

\begin{figure}[ht]
    \centering
    \includegraphics[width=0.50\textwidth]
    {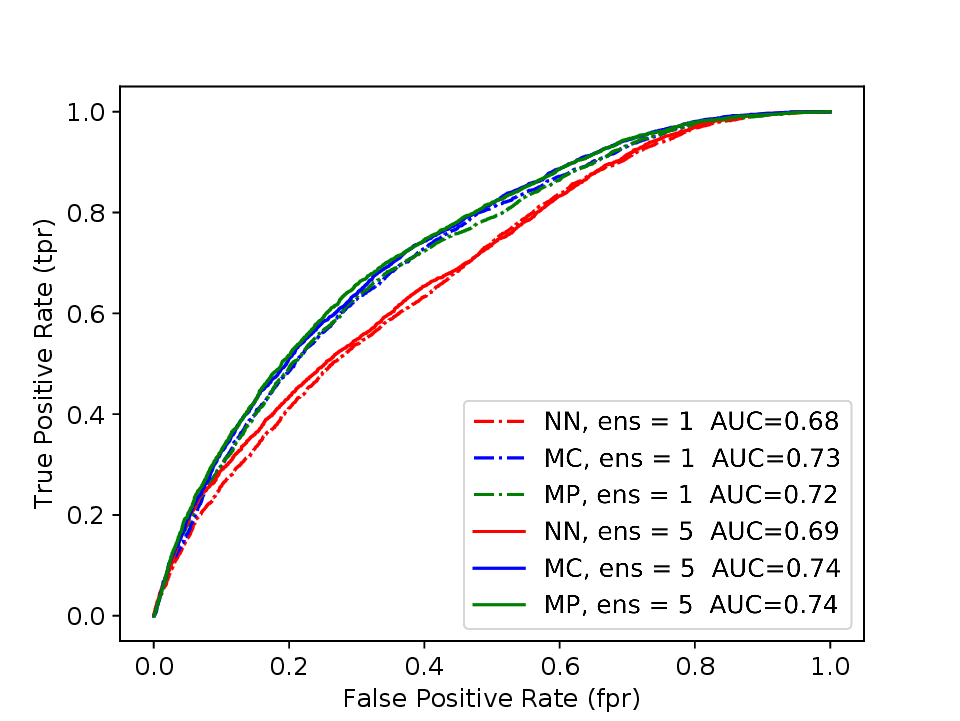}
    \caption{ROC curves for an OOD classification based on the entropy for CIFAR10 with five OOD classes. The different ROC curves correspond to the entropy values from NN, MC and MP for a single network (ens=1) and an network ensemble of five networks (ens=5) 
    }
    \label{fig:roc}
\end{figure}

The entropy, thus, can be seen as a continuous score to distinguish IND from OOD examples. With such a score, we can cast the OOD detection as a binary classification and do a ROC analysis. The ROC curve corresponding to the histograms in \ref{fig:ent_hist} is shown in figure \ref{fig:roc} with dashed lines. In addition the results for an ensemble of five trained models (ens=5) are shown, where each model was trained with different random initialization, and the predicted probabilities of the five models have been averaged. In \cite{hendrycks2016}, the area under the receiver operation curve (AUC) is suggested as an measure for the ability of an OOD detection method, as reported in the legend.

\begin{figure}[ht]
    \centering
    \includegraphics[width=0.45\textwidth]
    {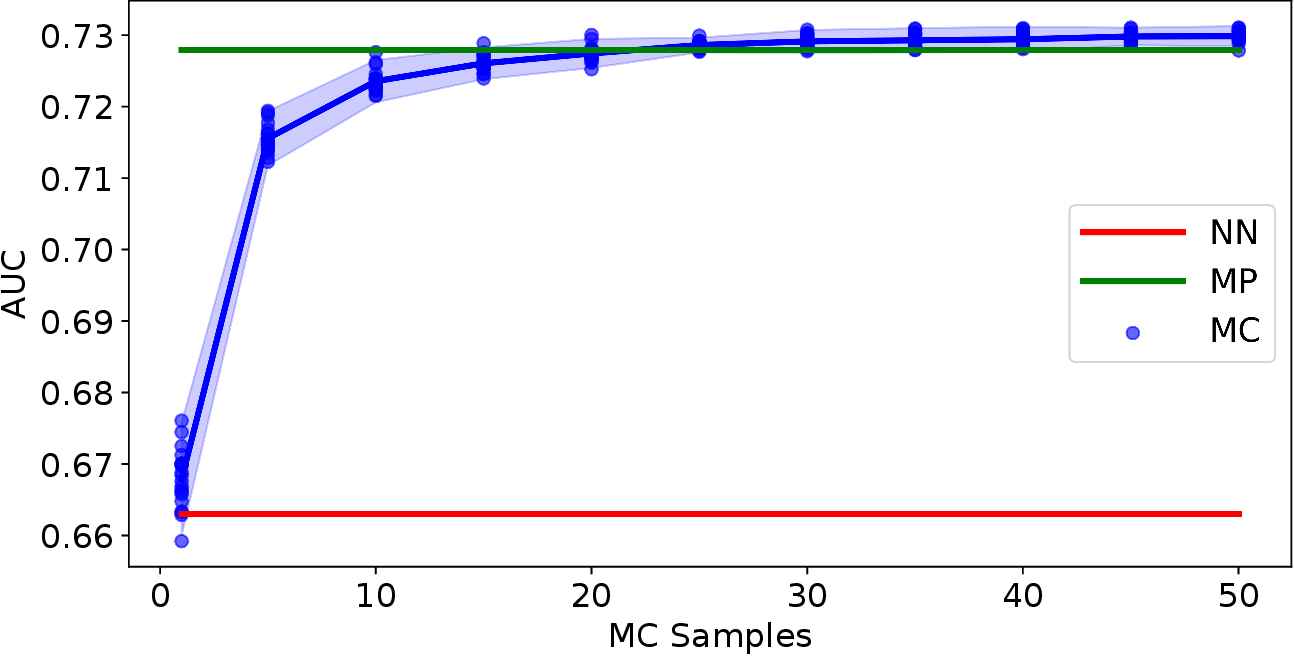}
    \caption{Dependence of AUC on the number of MC samples. Shown are the AUC values corresponding to the OOD classification experiment achieved with different models: a non-Bayesian NN (black horizontal line), our MP approach (red horizontal line), and an MC approach (blue). The MC approach was performed with different numbers of MC Samples $T$. Dotted MC results belong to 20 iterations, solid line indicates the average values.
    }
    \label{fig:auc}
\end{figure}

Finally, we investigate how many MC samples are required to achieve AUC values similar to the AUC values from the sampling-free MP approximation (see figure \ref{fig:auc}) indicating that our MP model can discriminate equally well between IND and OOD samples as the MC model with a certain number of sampling steps. 

To get a feeling for the reproducibility, we repeated the experiment 20 times. Figure \ref{fig:auc} shows that, with an increasing number of MC forward passes, the AUC of the MC method gradually improves and outperforms an NN. After $\approx$ 20 MC forward passes the AUC of the MC method is comparable to the AUC that we achieve with our MP method. At about $T=30$ the MC method slightly surpasses the performance of the MP approximation.

\begin{figure}[ht]
    \centering
    \includegraphics[width=0.50\textwidth]
    {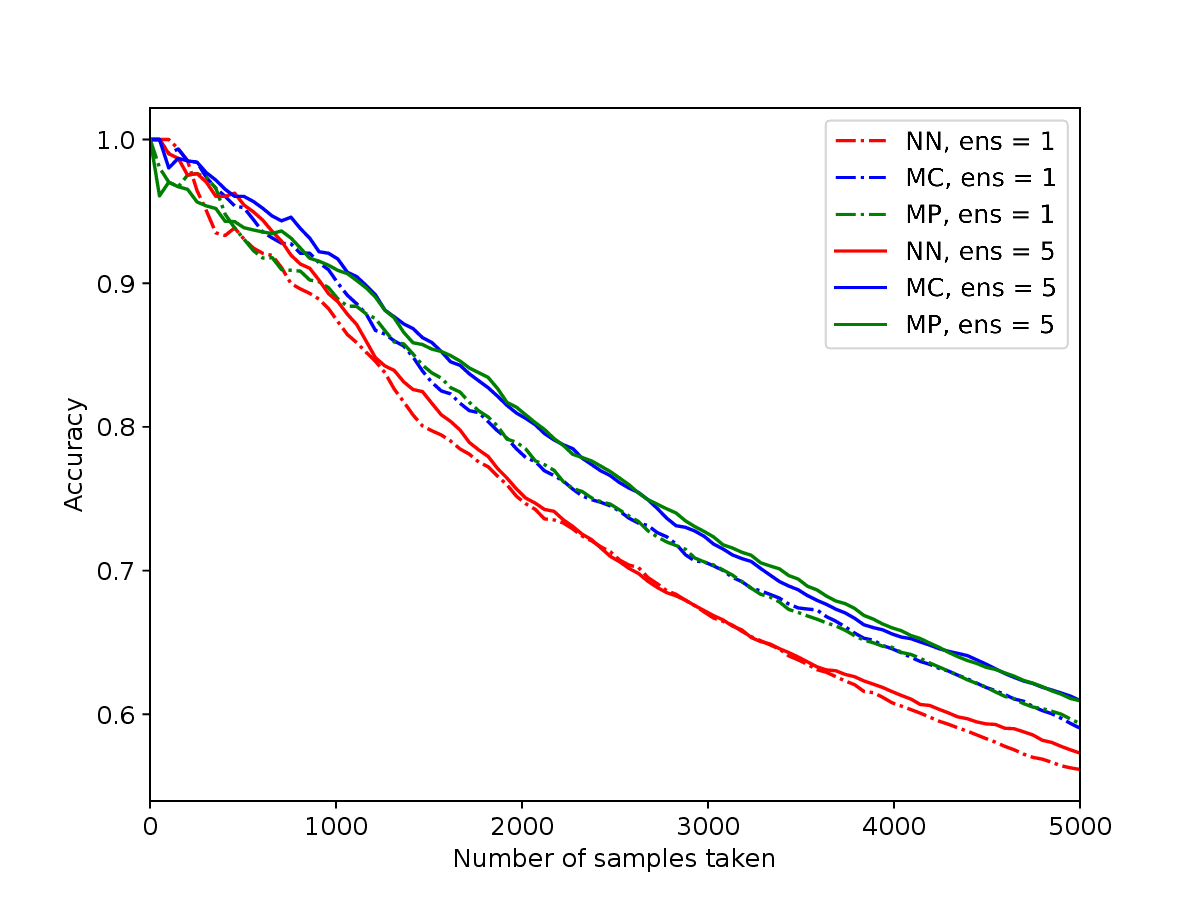}
    \caption{Filter experiment results where the entropy is used to filter out uncertain predictions. From left to right descending cutoffs of the entropy are used leading to an increasing number of samples that pass the less stringent uncertainty cutoff and a decreasing accuracy. The legend indicates the model type used: NN, MC, and MP for a single network (ens=1) and an ensemble of five networks (ens=5) 
    }
    \label{fig:filter}
\end{figure}

Finally, we compare the performance of our approximation to the MC method with a filter experiment relevant in practice. We, therefore, sort the 10 000 test examples according to their entropy value and then quantify for each threshold of the entropy the accuracy of the subset of test examples with entropy values below this threshold. In case of a useful uncertainty measure (low entropy values indicate reliable predictions), we expect better accuracies for smaller subsets corresponding to test examples with lower entropy values. With our MP method (see dotted lines in figure \ref{fig:filter}) we achieve a performance very similar to the MC approach, and both approaches are clearly superior to the simple NN. Also shown in the figure are the results for an ensemble of five members (ens=5). The beneficial effect of ensembling for MC and MP becomes evident but is interestingly less pronounced in case of the NN. 
%%%

	\section{Discussion}
\label{sec:discussion}
With our MP approach we have introduced an approximation to MC dropout that requires no sampling but instead propagates the expectation and the variance of the signal through the network. This results in a time saving by a factor that approximately corresponds to the number of MC runs. Further, the proposed approach can be applied to a traditional NN without re-training if the NN has been trained with standard dropout beforehand.

For the regression settings, we have shown that our MP approach approximates precisely the expectation and variance of the predictive distribution achieved by MC dropout. Also the achieved prediction performance in terms of RMSE and NLL do not show significant differences when using MC dropout or our MP approach. 

For the classification setting, we compared our approximation with the MC method. Specifically, we found no major difference between MC and MP in terms of the probabilities for the different classes and the derived entropy in both the IND and the OOD class setting. This suggests that the MP approximation can be used in all settings that require a precise quantification of uncertainty. To verify this assumption, we demonstrated that the beneficial effect of MC dropout remains preserved in the MP approach. This showed the ability of MC and MP models to improve the accuracy in the IND setting, to detect OOD examples, and to allow for filtering out wrongly classified images. We also demonstrated that the MP approach can be used in conjunction with ensembling.

Hence, our presented MP approach opens the door to include uncertainty information in real-time applications or to use the saved computing time for deep ensembling approaches leading to an additional boost in performance.

	\section{Acknowledgements}
\label{sec:acknowledgements}

We are very grateful to Elektrobit Automotive GmbH for supporting this research work. Further, part of the work has been founded by the Federal Ministry of	Education	and	Research of	Germany	(BMBF) in the project DeepDoubt (grant no. 01IS19083A).

	\bibliographystyle{unsrt}      % basic style, author-year citations
	%\bibliographystyle{ieee_fullname}
	
	\bibliography{mp_arxiv}
	
\end{document}